\documentclass{article}
\usepackage{array}
\usepackage{lineno,hyperref}
\usepackage{placeins}
\usepackage{moresize}
\usepackage{rotating}
\usepackage{graphicx}%
\usepackage{multirow}%
\usepackage{amsmath,amssymb,amsfonts}%
\usepackage{amsthm}%
\usepackage{mathrsfs}%
\usepackage[title]{appendix}%
\usepackage{xcolor}%
\usepackage{textcomp}%
\usepackage{manyfoot}%
\usepackage{booktabs}%
\usepackage{algorithmicx}%
\usepackage{algpseudocode}%
\usepackage{listings}%
\usepackage{tabularx}
\usepackage{comment}
\usepackage{adjustbox}
\usepackage{float}
\usepackage{tikz}
\usetikzlibrary{trees}

\usepackage{placeins}
\newtheorem{Definition}{Definition}

\usepackage{graphicx}

\usepackage{url}
\usepackage{hyperref}
\usepackage{subcaption}
\usepackage[justification=centering]{caption}
\usepackage[font=small,skip=0pt]{caption}
\usepackage{amsmath,amsfonts,amssymb}
\usepackage{multirow}
\usepackage{enumitem}
\usepackage{tabularx}
\usepackage[onelanguage,boxed,linesnumbered]{algorithm2e}

\usepackage{amsthm}

\date{}
\title{Hyperbolic Graph
Embeddings: a Survey and an Evaluation on Anomaly Detection}
\author{Souhail Abdelmouaiz SADAT$^1$
Mohamed Yacine TOUAHRIA MILIANI$^1$ \\
Khadidja HAB EL HAMES$^2$
 Hamida SEBA$^2$
 Mohammed HADDAD$^2$\\\\
$^1$\'{E}cole nationale Sup\'{e}rieure d'Informatique, Alger, Alg\'{e}rie.\\
\{js\_sadat, jm\_touahriamiliani\}@esi.dz  \\
$^2$Universit\'{e} de Lyon, CNRS, Universit\'{e} Lyon 1,\\
LIRIS, UMR5205, F-69622 Lyon, France.\\
\{hamida.seba, khadidja.hab-el-hames,  mohammed.hadded\}@univ-lyon1.fr}

\begin{document}
\maketitle
\date{}

\begin{abstract}
This survey reviews hyperbolic graph embedding models, and evaluate them on anomaly detection, highlighting their advantages over Euclidean methods in capturing complex structures. Evaluating models like \textit{HGCAE}, \textit{\(\mathcal{P}\)-VAE}, and \textit{HGCN} demonstrates high performance, with \textit{\(\mathcal{P}\)-VAE} achieving an F1-score of 94\% on the \textit{Elliptic} dataset and \textit{HGCAE} scoring 80\% on \textit{Cora}. In contrast, Euclidean methods like \textit{DOMINANT} and \textit{GraphSage} struggle with complex data. The study emphasizes the potential of hyperbolic spaces for improving anomaly detection, and provides an open-source library to foster further research in this field.
\end{abstract}


\section{Introduction}
\label{intro}

\quad In the era of digital transformation, the rapid increase in data complexity and volume has heightened the need for advanced anomaly detection techniques. Anomaly detection is essential across various domains, including cybersecurity, finance, and fraud detection, where it involves identifying unusual patterns that could indicate significant issues or potential threats.
Traditional anomaly detection methods often fall short when dealing with the growing complexity and scale of modern data. Conventional techniques, like classification and clustering, are generally designed for tabular data and struggle with the intricacies of more complex structures. Graph-based anomaly detection has emerged as a promising approach due to its ability to model intricate relationships within data. By representing data as graphs, where nodes denote entities and edges represent their interactions, this method captures complex patterns that might be missed by traditional techniques \cite{chandola2009anomaly}. Graph embeddings, which transform graph data into lower-dimensional vector spaces, further enhance anomaly detection by preserving the structural information and revealing subtle patterns indicative of anomalies \cite{hamilton2017inductive}.
A significant advancement in this field is the use of hyperbolic space for graph embeddings. Hyperbolic space, characterized by its constant negative curvature, provides a powerful framework for modeling hierarchical and complex relationships present in real-world data \cite{chami2019hyperbolic}. Unlike \textit{Euclidean} space, which can distort hierarchical structures, hyperbolic space allows for more accurate representation of these relationships, leading to improved performance in anomaly detection tasks 
\cite{tian2020survey}. Existing surveys have provided valuable insights into various aspects of hyperbolic geometry in machine learning and graph-based methods, yet there remains a lack of comprehensive syntheses and algorithmic analyses of hyperbolic graph embedding techniques. For instance, \cite{9658224}  provides a general survey on hyperbolic deep neural networks (\textit{HDNNs}), exploring their architectures and applications across diverse domains, while \cite{yang2023hyperbolicgraphneuralnetworks} focuses specifically on hyperbolic graph neural networks (\textit{HGNNs}), unifying existing approaches into a general framework and summarizing their key components and applications. 
However, while these studies provide valuable context, None of these surveys have thoroughly examined the feasibility of these methods on a specific use case to establish a well-founded comparison. Moreover, none of them implement these methods for experimental benchmarking. In our survey, not only do we review and classify existing methods, but we also provide a library that implements the most significant approaches, along with a framework that allows for testing their effectiveness in the task of anomaly detection, a classification problem. By bridging this gap, our work offers both a theoretical synthesis and a practical evaluation, facilitating a more rigorous assessment of these methods in real-world applications.
Our contributions include:
\begin{itemize}
       \item \textit{Detailed Review and Analysis : } We present a thorough review a of existing hyperbolic graph embedding techniques.
     \item \textit{Exploration on anomaly detection:} We investigate the application of hyperbolic graph embeddings on anomaly detection, highlighting their advantages over traditional methods by evaluating them on established datasets.
    \item \textit{Development of an Open-Source Library :} We present an open-source library that includes the main hyperbolic graph embedding methods and datasets \url{https://gitlab.liris.cnrs.fr/gladis/ghypeddings}.

\end{itemize}

The remainder of this paper is structured as follows: Section \ref{section: preleminaries} provides the necessary background on graph embedding and highlighting the main methods in the \textit{Euclidean} space. Section \ref{section:taxonomy} introduces the hyperbolic space covering key concepts in differential geometry, and hyperbolic geometry graph and its key concepts.  It also proposes a comprehensive taxonomy of hyperbolic graph embedding models, categorizing the existing techniques into traditional methods and deep learning-based methods. 
Section \ref{section:solution} outlines the methodology we followed for using hyperbolic embedding for the anomaly detection task. We also introduce the Ghypeddings library we constructed for this purpose. Section \ref{section:expermient} describes the experimental setup, including datasets, evaluation metrics, and presents the experimental results, discussing the key findings. Finally, Section \ref{section:conclusion} concludes the paper and offers insights into potential future research directions.

\section{Graph Embedding in the \textit{Euclidean} Space}
\label{section: preleminaries}

In this section, we define the concepts of graphs and graph embedding. Then we present the commonly used techniques of graph embedding in the \textit{Euclidean} space.

Graphs are powerful data structures used to model relationships among entities in various domains. However, their irregular structure presents challenges for conventional machine learning models designed to process tabular data. Graph embeddings address this issue by mapping graph data into a low-dimensional, continuous vector space that captures structural and attribute-based information. This transformation enables machine learning models to effectively utilize graph data for various tasks.\\
Formally, let $G = (V, E)$ represent a graph, where $V$ is the set of nodes and $E$ is the set of edges. A \textit{graph embedding} is a mapping $f: V \rightarrow \mathbb{R}^d$, where each node $v \in V$ is represented as a $d$-dimensional vector in \textit{Euclidean} space. The objective is to ensure that geometric relations in the embedding space reflect structural or attribute-based similarities in the original graph. These embeddings are vital for tasks such as node classification, link prediction, graph clustering, and visualization.\\
Graph embedding methods can be categorized based on their tasks and objectives. Supervised tasks involve predicting outputs using graph structure and node labels, such as node and graph classification. In contrast, unsupervised tasks leverage the graph's inherent structure for self-supervision, enabling applications like link prediction, graph reconstruction, and clustering \cite{JMLR:v23:20-852}.

Graph embedding methods often operate at the node level, with edge and graph embeddings derived from node embeddings. Most techniques employ the \textit{Euclidean} space $\mathbb{R}^n$ as the embedding space. The study of graph embeddings dates back to the early 2000s, when they were explored for dimensionality reduction tasks \cite{cai2018comprehensive}. In this context, a graph is typically constructed from high-dimensional data points by connecting each point to its top $k$ nearest neighbors. Pairwise node similarity matrices are then calculated, and nodes are mapped into a lower-dimensional space to preserve these pairwise relationships. For instance, \textit{Isomap} \cite{tenenbaum2000global} computes shortest path distances on the graph and applies multidimensional scaling (MDS) \cite{kruskal1964multidimensional}, which reconstructs a set of points from a matrix of pairwise distances. 
Numerous \textit{Euclidean} graph embedding methods have since been proposed
 \cite{hamilton2017Representation}. These methods are genrally classified into three primary algorithmic categories:

\paragraph{\textbf{\underline{1- Matrix Factorization-Based Methods:}}} These methods decompose a similarity matrix into lower-dimensional matrices to learn node embeddings that preserve node relationships. \textit{Graph Factorization} factorizes the adjacency matrix \( A \) to learn node embeddings such that the inner product of embeddings approximates the graph's connectivity \cite{ahmed2013distributed}.

\paragraph{\textbf{\underline{2- Random Walk-Based Methods:}}} These methods generate stochastic sequences (random walks) over the graph and use these sequences to learn node embeddings. These methods are flexible, as they can model various types of node proximities by adjusting the random walk behavior.

\paragraph{\textbf{\underline{3- Deep Learning-Based Methods:}}}  Deep learning techniques, specifically \textit{Graph Neural Networks} (\textit{GNNs}), have become central to graph embedding. \textit{GNNs} can learn node embeddings by aggregating information from a node’s neighborhood iteratively, allowing the model to capture complex structural dependencies within the graph.\\
\textit{Graph Convolutional Networks (GCNs)} \cite{kipf2017semi} use a convolutional operation over the graph to aggregate information from a node's neighbors and update its embedding. 
\textit{GraphSAGE} (Graph Sample and Aggregation) \cite{hamilton2017inductive}, unlike \textit{GCNs}, which require the full graph to be stored in memory, performs inductive learning by sampling a fixed-size neighborhood for each node. This allows the model to scale to large graphs. 
\textit{Graph Attention Networks (GATs)} \cite{velickovic2018graph} introduce a self-attention mechanism to \textit{GNNs}. In \textit{GATs}, each node learns to assign attention coefficients to its neighbors, which indicate the importance of each neighbor's information. 
 \textit{Graph Autoencoders (GAEs)} models \cite{kipf2016variational} adapt the autoencoder framework to graph-structured data. The encoder learns node embeddings, while the decoder reconstructs the graph (e.g., adjacency matrix) from the embeddings. \textit{Variational Graph Autoencoders (VGAEs)} introduce probabilistic elements to the framework, allowing for uncertainty modeling in the embeddings.

\section{Hyperbolic Graph Embedding}
\label{section:taxonomy}
In this section, we move to the hyperbolic space. We first introduce the fundamental concepts of differential geometry. Then, we provide a detailed overview of the hyperbolic space and its commonly used models. Finally, we thoroughly review existing hyperbolic graph embedding techniques.

\subsection{Differential Geometry}
\label{subsec:diff-geo}

\begin{Definition} [Topological Space] Let $X$ be a set, and $\tau$ be a collection of subsets of $X$. The pair $(X, \tau)$ is said to be a topological space, and $\tau$ is called a topology on $X$ if it satisfies three 
axioms: 
 (1) The empty set $\varnothing$ and $X$ belong to $\tau$.
   (2) Any finite or infinite union of members of $\tau$ belongs to $\tau$.
     (3) Any finite intersection of members of $\tau$ belongs to $\tau$.
\end{Definition}

\begin{Definition}[Manifold] A manifold $\mathcal{M}$ of dimension $n$ is a \textit{topological space} in which the neighborhood of any point can be locally approximated by \textit{Euclidean} space. For every point $p \in \mathcal{M}$, there exists a pair $(U, \varphi)$ called a \textit{coordinate chart}, where $U$ is a neighborhood of $p$, and $\varphi$ is a \textit{homeomorphism} (a continuous bijection with a continuous inverse) between $U$ and an open subset of $\mathbb{R}^n$. Manifolds are a higher-dimensional generalization of surfaces.
\end{Definition}

\begin{Definition}[Smooth Manifold] Given a \textit{manifold} $\mathcal{M}$ equipped with a collection of \textit{coordinate charts} that cover $\mathcal{M}$, we say that $\mathcal{M}$ is \textit{smooth} if, for every two \textit{coordinate charts} $(U_\alpha, \varphi_\alpha)$ and $(U_\beta, \varphi_\beta)$ whose domains intersect, the map $\varphi_\beta \circ \varphi_\alpha^{-1}$ and its inverse are $C^\infty$ (infinitely differentiable). This property allows for the application of differential calculus on $\mathcal{M}$.
\end{Definition}

\begin{Definition}[Tangent Space] Let $\mathcal{M}$ be an $n$-dimensional smooth manifold and $p \in \mathcal{M}$. The \textit{tangent space} of $\mathcal{M}$ at $p$, denoted by $\mathcal{T}_p\mathcal{M}$, is the collection of all possible \textit{tangent vectors} at $p$. Intuitively, $\mathcal{T}_p\mathcal{M}$ is the \textit{vector space} that best approximates the manifold in a neighborhood of $p$.
\end{Definition}

\begin{Definition}[Riemannian Manifold] A \textit{Riemannian manifold} $(\mathcal{M}, g)$ is a smooth manifold $\mathcal{M}$ equipped with a \textit{Riemannian metric} $g$, which is a smoothly varying inner product $g_p : \mathcal{T}_p\mathcal{M} \times \mathcal{T}_p\mathcal{M} \rightarrow \mathbb{R}$ defined on the tangent space at each point $p \in \mathcal{M}$. The $n \times n$ symmetric positive-definite matrix $G(p)$ that satisfies $g_p(u,v) = u^\top G(p) v$ is called the \textit{metric tensor} of $\mathcal{M}$ at $p$.
\end{Definition}

\begin{Definition} [Geodesics] Geodesics generalize straight lines in \textit{Euclidean} space to curved spaces. Formally, a \textit{geodesic} is a constant-speed curve that minimizes the distance between two points locally. The \textit{distance} between two points on the manifold is the length of the geodesic connecting them.
\end{Definition}

\begin{Definition}[Exponential Map] On a Riemannian manifold $(\mathcal{M}, g)$, the \textit{exponential map} $\exp_p : \mathcal{T}_p\mathcal{M} \rightarrow \mathcal{M}$ projects a tangent vector $v \in \mathcal{T}_p\mathcal{M}$ to the point on $\mathcal{M}$ reached by following the geodesic starting at $p$ in the direction of $v$.
\end{Definition}

\begin{Definition}[Logarithmic Map] The \textit{logarithmic map} $\log_p : \mathcal{M} \rightarrow \mathcal{T}_p\mathcal{M}$ is the inverse of the exponential map. It projects points from the manifold $\mathcal{M}$ back to the tangent space $\mathcal{T}_p\mathcal{M}$.
\end{Definition}


\begin{Definition}[Gaussian Curvature] The Gaussian curvature of a surface $S$ at a point $p \in S$ measures the deviation of $S$ from being flat at $p$. It is a signed value where positive curvature corresponds to a spherical surface, negative curvature to a saddle-shaped surface, and zero curvature to a flat plane.
\end{Definition}

\begin{Definition}[Sectional curvature] Let $\mathcal{M}$ be a Riemannian manifold and $\mathcal{T}_p\mathcal{M}$ the tangent space of $\mathcal{T}$ at $p \in \mathcal{M}$. The sectional curvature of $\mathcal{M}$ at $p$ associated with a $2-$dimensional plane $\Pi_p \subset \mathcal{T}_p\mathcal{M}$ is defined as the Gaussian curvature of the surface which has the plane $\Pi_p$ as a tangent plane at $p$. $\mathcal{M}$ is said to have \textit{constant sectional curvature} if its sectional curvature at every point and for all $2-$dimensional planes $\Pi_p \subset \mathcal{T}_p\mathcal{M}$ is the same. The notion of sectional curvature is a generalization of Gaussian curvature for high-dimensional Riemannian manifolds.
\end{Definition}

\subsection{Hyperbolic Geometry}
\label{subsec:hyp-geo}
In this section, we briefly review the fundamental concepts of hyperbolic geometry. For a more detailed examination, refer to \cite{lee_introduction_nodate}.\\
\quad Hyperbolic geometry is a non-\textit{Euclidean} geometry characterized by constant negative curvature, which contrasts with \textit{Euclidean} geometry where the parallel postulate holds. This results in unique properties for lines, angles, and distances.\\
The \textit{hyperbolic space} of $n-$dimensions is defined as a complete\footnote{A Riemannian manifold is 	\textbf{complete} if every geodesic can be extended indefinitely.} and simply connected\footnote{A manifold is \textbf{simply connected} if it has no holes, meaning every loop can be contracted to a point.} $n-$dimensional Riemannian manifold with \textit{constant negative sectional curvature}. This space can be represented using five well-known models. Each model is defined on a different subset of the real vector space, called its domain, and has its own tensor metric, geodesics, distance function, etc \cite{cannon1997hyperbolic}.
We focus on three widely studied models of hyperbolic space, the  \textit{Lorentz} model, the \textit{Poincaré} ball model, and the Klein model, each offering a different perspective but being mathematically equivalent. These models are defined as follows:
\paragraph{\textbf{ \textit{Lorentz} Model}} Also known as the Hyperboloid model or the
Minkowski model. The  \textit{Lorentz} model of $n-$dimensional hyperbolic space corresponds to the Riemannian manifold $(\mathbb{L}^n,g^\mathbb{L})$ where $\mathbb{L}^n$ is given as the upper sheet of a two-sheeted $n-$dimensional hyperbola, that is
$\mathbb{L}^n = \{x=(x_1,x_2,\hdots,x_{n+1} ) \in \mathbb{R}^{n+1}: \langle x,x \rangle_\mathbb{L}=-1, x_1>0\}$,
in which the $\langle x,x \rangle_\mathbb{L}$ denotes the  \textit{Lorentzian} inner product which is defined for every $x,y \in \mathbb{L}^n$ as
\begin{displaymath}
\langle x,y \rangle_\mathbb{L}=-x_1y_1 + \sum_{i=1}^{n+1} x_i y_i .
\end{displaymath}
Its metric tensor $g^\mathbb{L}_x$ is a diagonal matrix of size $n+1$ such that each main diagonal element is equal to 1, except for the first element being -1, i.e.,
$g^\mathbb{L}_x = \text{diag}(-1,1,\hdots,1)$.
The induced distance on the  \textit{Lorentz} manifold between two points $x,y \in \mathbb{L}^n$ is by
%
$d_\mathbb{L}(x,y)=\text{arcosh}(-\langle x,y \rangle_\mathbb{L})$.
\paragraph{\textbf{Poincaré ball model}} The n$-$dimensional \textit{Poincaré} ball model of hyperbolic space is defined as the Riemannian manifold $(\mathbb{B}^n,g^\mathbb{B})$ where $\mathbb{B}^n$ is an open unit ball defined as
$\mathbb{B}^n=\{x \in \mathbb{R}^n: \|x\|< 1\}$,
in which $\|.\|$ denotes the \textit{Euclidean} norm. The metric tensor $g^\mathbb{B}_x$ is given for any $x \in \mathbb{B}^n$ as follows:
\begin{displaymath}
g^\mathbb{B}_x = \lambda_x^2 I_n \,, \quad \text{where} \ \lambda_x = \frac{2}{1-\|x\|^2} \,.
\end{displaymath}
The $2-$dimensional version of the \textit{Poincaré} ball model is known as the \textit{Poincaré} disk. The associated distance function between two points $x,y \in \mathbb{B}^n$ is computed as:
\begin{displaymath}
d_\mathbb{B}(x,y)=\text{arcosh}\left(1+2\frac{\|x-y\|^2}{(1-\|x\|^2 )(1-\|y\|^2 )}\right) .
\end{displaymath}

\paragraph{\textbf{Klein model}} Also known as the Beltrami–Klein model or the projective model, The Klein model of $n-$dimensional hyperbolic space is the Riemannian manifold $(\mathbb{K}^n,g^\mathbb{K})$ where $\mathbb{K}^n$ is the open unit ball given by
$\mathbb{K}^n=\{x \in \mathbb{R}^n: \|x\|< 1\}$,
and $g^\mathbb{K}_x$ is the metric tensor defined for any $x \in \mathbb{K}^n$ as
\begin{displaymath}
g^\mathbb{K}_x = \frac{I_n}{1-\|x\|^2} + \frac{x.x^\top}{(1-\|x\|^2)^2} \,.
\end{displaymath}

\subsection{Taxonomy of Hyperbolic Graph Embedding Methods}
\begin{figure}[!th]
\centering
\begin{tikzpicture}[every node/.style={font=\scriptsize}]
    \def\xsep{0.1} 
    \def\ysep{0.3} 
   \def\ydeb{5};

    \def\xline{0.1}; 

    \def\ytop{0*\ysep+\ydeb};
    \def\ybottom{-11*\ysep+\ydeb};
    \draw (\xline,\ybottom) -- (\xline,\ytop);
         \draw (0.1,-5.5*\ysep+\ydeb) -- (0.3,-5.5*\ysep+\ydeb);
        \node at (1.1,-5*\ysep+\ydeb) {Convolutional};
        \node at (1.1,-6*\ysep+\ydeb) { GNNs};
    \node[anchor=east] at (0,\ydeb) {HGNN \cite{liu2019hyperbolic}};
    \node[anchor=east] at (0,-\ysep+\ydeb) {HGCN \cite{chami2019hyperbolic}};
    \node[anchor=east] at (0,-2*\ysep+\ydeb) {k-GCN \cite{pmlr-v119-bachmann20a}};
    \node[anchor=east] at (0,-3*\ysep+\ydeb) {HAT \cite{zhang2021hyperbolic}};
    \node[anchor=east] at (0,-4*\ysep+\ydeb) {GIL \cite{NEURIPS2020_551fdbb8}};
    \node[anchor=east] at (0,-5*\ysep+\ydeb) {Q-GCN \cite{NEURIPS2022_16c628ab}};
    \node[anchor=east] at (0,-6*\ysep+\ydeb) {ACE-HGNN \cite{9679192}};
    \node[anchor=east] at (0,-7*\ysep+\ydeb) {NHGCN \cite{fan2022nested}};
    \node[anchor=east] at (0,-8*\ysep+\ydeb) {kHGCN \cite{kHGCN} };
    \node[anchor=east] at (0,-9*\ysep+\ydeb) {FMGNN \cite{deng2023fmgnnfusedmanifoldgraph}};
    \node[anchor=east] at (0,-10*\ysep+\ydeb) {H2H-GCN \cite{9578363}};
    \node[anchor=east] at (0,-11*\ysep+\ydeb) {LGCN \cite{zhang2021lorentzian}};



    \foreach \i in {0,1,2,3,4,5,6,7,8,9,10,11} {
        \draw (\xline,-\i*\ysep+\ydeb) -- (\xline-\xsep,-\i*\ysep+\ydeb);
    }

    \def\ytb{-14*\ysep+\ydeb}
    \def\ybb{-18*\ysep+\ydeb}
    \draw (\xline,\ytb) -- (\xline,\ybb);
    \draw (0.1,-16*\ysep+\ydeb) -- (0.3,-16*\ysep+\ydeb);
    \node at (1.1,-16*\ysep+\ydeb) {Graph };
    \node at (1.1,-17*\ysep+\ydeb) { Autoencoders};
    \node[anchor=east] at (0,-14*\ysep+\ydeb) {$\mathcal{P}$-VAE \cite{mathieu2019continuous}};
    \node[anchor=east] at (0,-15*\ysep+\ydeb) {HGCAE \cite{park2021unsupervised}};
    \node[anchor=east] at (0,-16*\ysep+\ydeb) {CCM-AAE \cite{GRATTAROLA2019105511}};
    \node[anchor=east] at (0,-17*\ysep+\ydeb) {PWA \cite{ovinnikov2019poincar}};
    \node[anchor=east] at (0,-18*\ysep+\ydeb) {HyperbolicNF \cite{bose2020latent}};

    \foreach \i in {14,15, 16, 17,18} {
        \draw (\xline,-\i*\ysep+\ydeb) -- (\xline-\xsep,-\i*\ysep+\ydeb);
    }

    \def\ytb{-21*\ysep+\ydeb}
    \def\ybb{-25*\ysep+\ydeb}
    \draw (\xline,\ytb) -- (\xline,\ybb);
    \draw (0.1,-23.5*\ysep+\ydeb) -- (0.3,-23.5*\ysep+\ydeb);
    \node at (1.2,-23*\ysep+\ydeb) {Spatio-temporal };
    \node at (1.1,-24*\ysep+\ydeb) {GNNs};
    \node[anchor=east] at (0,-21*\ysep+\ydeb) {HTGN \cite{DBLP:journals/corr/abs-2107-03767}};
    \node[anchor=east] at (0,-22*\ysep+\ydeb) {HGWaveNet \cite{Bai_2023}};
    \node[anchor=east] at (0,-23*\ysep+\ydeb) {ST-GCN \cite{10.1145/3394171.3413910}};
    \node[anchor=east] at (0,-24*\ysep+\ydeb) {HVGNN \cite{sun2021hyperbolic}};
    \node[anchor=east] at (0,-25*\ysep+\ydeb) {DHGAT \cite{LI2024127038}};
    \foreach \i in {21,22,23,24,25} {
        \draw (\xline,-\i*\ysep+\ydeb) -- (\xline-\xsep,-\i*\ysep+\ydeb);
    }
  \def\xlined{7} 
    %
    \def\ytda{0*\ysep+\ydeb}
    \def\ybda{-11*\ysep+\ydeb}  
    \draw (\xlined-0.1,\ytda) -- (\xlined-0.1,\ybda);
    \draw (\xlined-0.3,-6*\ysep+\ydeb) -- (\xlined-0.1,-6*\ysep+\ydeb);
    \node at (\xlined-0.9,-6*\ysep+\ydeb) {Direct};
    \node at (\xlined-1,-7*\ysep+\ydeb) {Optimization};

    \node[anchor=west] at (\xlined,\ydeb) {Poincaré};
    \node[anchor=west] at (\xlined,-\ysep+\ydeb) {Embedding \cite{nickel2017poincare}};
    \node[anchor=west] at (\xlined,-2*\ysep+\ydeb) {Lorentz };
    \node[anchor=west] at (\xlined,-3*\ysep+\ydeb) {embedding \cite{nickel2018learning}};
    \node[anchor=west] at (\xlined,-4*\ysep+\ydeb) {Entailment};
    \node[anchor=west] at (\xlined,-5*\ysep+\ydeb) {Cones \cite{ganea2018hyperbolic}};
    \node[anchor=west] at (\xlined,-6*\ysep+\ydeb) {Lorentzian};
    \node[anchor=west] at (\xlined,-7*\ysep+\ydeb) {distance \cite{law2019lorentzian}};
    \node[anchor=west] at (\xlined,-8*\ysep+\ydeb) {Disk};
    \node[anchor=west] at (\xlined,-9*\ysep+\ydeb) {embedding \cite{suzuki2019hyperbolic}};
    \node[anchor=west] at (\xlined,-10*\ysep+\ydeb) {Tiling \cite{NEURIPS2019_82c25591}};
    \node[anchor=west] at (\xlined,-11*\ysep+\ydeb) {Hyperbolic random};
    \node[anchor=west] at (\xlined,-12*\ysep+\ydeb) {walk 
    \cite{wang2019hyperbolic}};
    \foreach \i in {0,2,4,6,8,10,11} {  
        \draw (\xlined,-\i*\ysep+\ydeb) -- (\xlined-\xsep,-\i*\ysep+\ydeb);
    }

    \def\ytdb{-15*\ysep+\ydeb}
    \def\ybdb{-16*\ysep+\ydeb}  
    \draw (\xlined-0.1,\ytdb) -- (\xlined-0.1,\ybdb);
    \draw (\xlined-0.3,-15.5*\ysep+\ydeb) -- (\xlined-0.1,-15.5*\ysep+\ydeb);
    \node at (\xlined-0.9,-15.5*\ysep+\ydeb) {Tree};
    \node at (\xlined-1,-16.5*\ysep+\ydeb) {Approximation};

    \node[anchor=west] at (\xlined,-15*\ysep+\ydeb) {shortest path 
    \cite{yim2023fitting}};
    \node[anchor=west] at (\xlined,-16*\ysep+\ydeb) {Combinatorial};
    \node[anchor=west] at (\xlined,-17*\ysep+\ydeb) {constructions \cite{sala2018representation}};

    \foreach \i in {15,16} {
        \draw (\xlined,-\i*\ysep+\ydeb) -- (\xlined-\xsep,-\i*\ysep+\ydeb);
    }

    \def\ytdb{-20*\ysep+\ydeb}
    \def\ybdb{-29*\ysep+\ydeb}
   \draw (\xlined-0.1,\ytdb) -- (\xlined-0.1,\ybdb);
    \draw (\xlined-0.3,-24*\ysep+\ydeb) -- (\xlined-0.1,-24*\ysep+\ydeb);
    \node at (\xlined-1,-24*\ysep+\ydeb) {Multi-};
     \node at (\xlined-1,-25*\ysep+\ydeb) {dimensional};
        \node at (\xlined-1,-26*\ysep+\ydeb) {scaling};
    \node[anchor=west] at (\xlined,-20*\ysep+\ydeb) {H-MDS \cite{walter2004h}};
    \node[anchor=west] at (\xlined,-21*\ysep+\ydeb) {PD-MDS \cite{cvetkovski2011multidimensional}};
    \node[anchor=west] at (\xlined,-22*\ysep+\ydeb) {Lorentzian};
    \node[anchor=west] at (\xlined,-23*\ysep+\ydeb) {MDS \cite{10.1371/journal.pone.0187301}};
    \node[anchor=west] at (\xlined,-24*\ysep+\ydeb) {h-MDS \cite{pmlr-v80-sala18a}};
    \node[anchor=west] at (\xlined,-25*\ysep+\ydeb) {hydra \cite{keller2020hydra}};
     \node[anchor=west] at (\xlined,-26*\ysep+\ydeb) {HDM \cite{tabaghi2020hyperbolic}};
     \node[anchor=west] at (\xlined,-27*\ysep+\ydeb) {PGA \cite{huckemann2010intrinsic}};
\node[anchor=west] at (\xlined,-28*\ysep+\ydeb) {Symmetric spaces \cite{pmlr-v139-lopez21a}};
\node[anchor=west] at (\xlined,-29*\ysep+\ydeb) {Manifold product \cite{gu2018learning}};
    \foreach \i in {20,21,22,24,25,26,27, 28,29} {
       \draw (\xlined,-\i*\ysep+\ydeb) -- (\xlined-\xsep,-\i*\ysep+\ydeb);
    }
     \def\ytdb{-32*\ysep+\ydeb}
    \def\ybdb{-36*\ysep+\ydeb}
    \draw (\xlined-0.1,\ytdb) -- (\xlined-0.1,\ybdb);
   \draw (\xlined-0.3,-34*\ysep+\ydeb) -- (\xlined-0.1,-34*\ysep+\ydeb);
    \node at (\xlined-1,-33.5*\ysep+\ydeb) {Network};
        \node  at (\xlined-1,-34.5*\ysep+\ydeb) {Model};
    \node[anchor=west] at (\xlined,-32*\ysep+\ydeb) {Community Structure \cite{WANG2016609}};
    \node[anchor=west] at (\xlined,-33*\ysep+\ydeb) {Common neighbors \cite{8314557}};
    \node[anchor=west] at (\xlined,-34*\ysep+\ydeb) {Laplacian \cite{alanis2016efficient}};
    \node[anchor=west] at (\xlined,-35*\ysep+\ydeb) {HyperMap \cite{papadopoulos2014network}};
    \node[anchor=west] at (\xlined,-36*\ysep+\ydeb) {HyperMap-CN \cite{papadopoulos2015network}};

%
    \foreach \i in {32,33,34,35,36} {
        \draw (\xlined,-\i*\ysep+\ydeb) -- (\xlined-\xsep,-\i*\ysep+\ydeb);
    }

\draw (2.3,6) -- (2.3,-23.5*\ysep+\ydeb);
\draw  (2.3,-5*\ysep+\ydeb) -- (2.1,-5*\ysep+\ydeb) ;
\draw  (2.3,-16*\ysep+\ydeb) -- (2.1,-16*\ysep+\ydeb);
\draw  (2.3,-23.5*\ysep+\ydeb) -- (2.1,-23.5*\ysep+\ydeb);

\draw (\xlined-2,6) -- (\xlined-2,-34*\ysep+\ydeb);
\draw (\xlined-2,-6*\ysep+\ydeb) -- (\xlined-1.8,-6*\ysep+\ydeb);	
\draw (\xlined-2,-15.5*\ysep+\ydeb) -- (\xlined-1.8,-15.5*\ysep+\ydeb);
\draw (\xlined-2,-24*\ysep+\ydeb) -- (\xlined-1.8,-24*\ysep+\ydeb);
\draw (\xlined-2,-34*\ysep+\ydeb) -- (\xlined-1.8,-34*\ysep+\ydeb);	

\node[fill=gray!20] at (3.5,7) {Hyperbolic Graph Embedding};
\node[fill=gray!20] at (2,6) {Deep learning methods};
\node[fill=gray!20] at (5,6) {Traditional methods};

\draw[->] (3.5,6.8) -- (2,6.2);	
\draw[->] (3.5,6.8) -- (5,6.2);

   \end{tikzpicture}
   \caption{Taxonomy of hyperbolic graph embedding methods.}
    \label{fig:taxonomy}
    \end{figure}
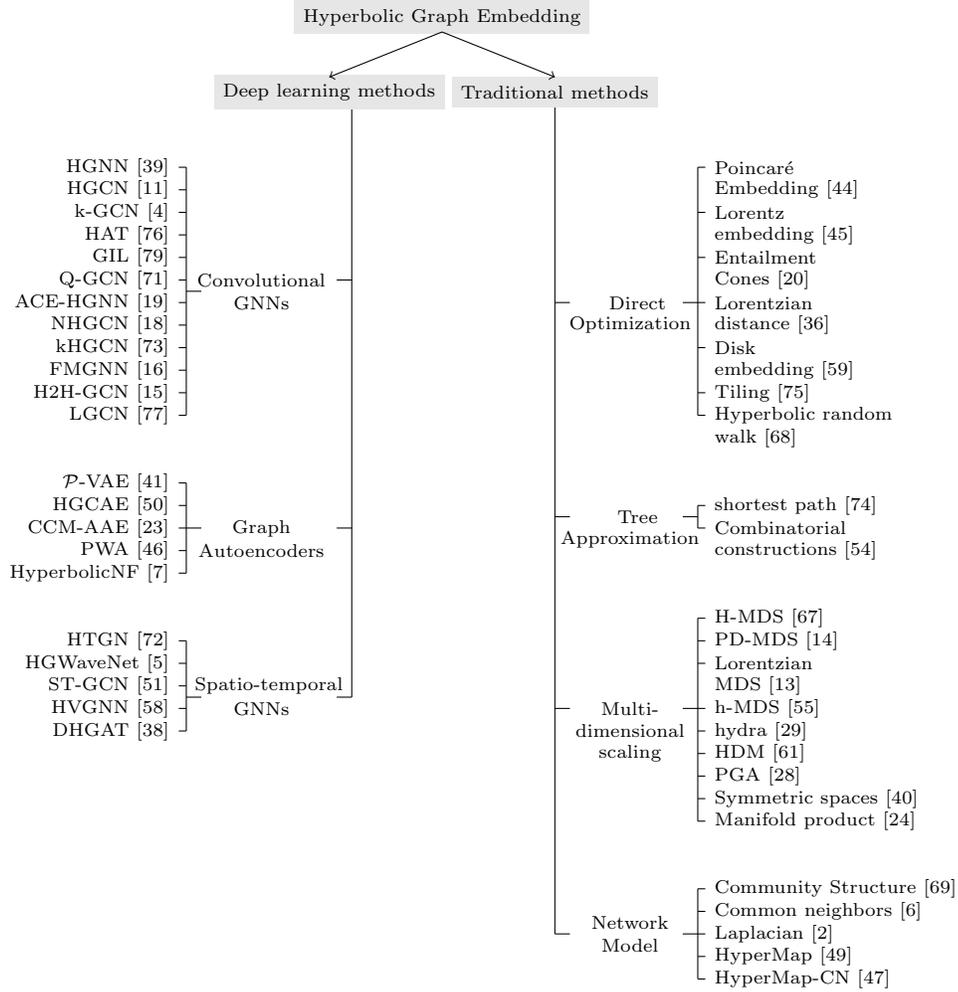

Several approaches have been proposed in the literature for embedding graphs into hyperbolic spaces, and these methods can be categorized based on various criteria. Key distinctions include whether the methods incorporate node features or focus solely on preserving graph structure, as well as the level of supervision. Unsupervised methods typically learn embeddings in two stages, while supervised methods optimize directly for specific tasks like node classification. Additionally, embeddings can be transductive, mapping nodes directly, or inductive, learning a function that can embed new nodes. Based on these distinctions, we propose a comprehensive taxonomy that categorizes state-of-the-art methods (see Figure \ref{fig:taxonomy}). 
In what follows, we describe the categories within our proposed taxonomy and present in detail the most representative methods for each category (that we evaluate in our experimental section) while describing briefly the other methods.

\noindent \underline{\textbf{{1- Traditional methods }}}
Traditional methods for embedding graphs into hyperbolic spaces can be categorized based on their approach and the type of information they utilize. These methods include direct optimization-based approaches, tree approximation-based techniques, multidimensional scaling (MDS)-based methods and network model dependent methods.
\begin{table}[t]
\centering
\caption{Summary of the main Traditional hyperbolic embedding approaches.}
\label{tab:embedding_comparison_transposed}
\footnotesize
\begin{tabular}{p{1.5cm} p{3.3cm} p{2.5cm} p{3cm}}
\toprule
{\textbf{Category}} &
{\textbf{\shortstack{Direct Optimization}}} &
{\textbf{\shortstack{Tree Approximation}}} &
{\textbf{\shortstack{Multidimensional Scaling}}} \\
\midrule
\textbf{Approach} & \textit{Poincaré} Embeddings \cite{nickel2017poincare} & Combinatorial Construction \cite{sala2018representation} & \textit{H-MDS} \cite{walter2004h} \\ 
\textbf{Objective} & Minimize task-specific loss & Embed graph as a weighted tree & Extend MDS to hyperbolic space \\ 
\textbf{Embedding Space} & \textit{Poincaré} Ball & \textit{Poincaré} Ball &  \textit{Lorentz} manifold \\ 
\textbf{Tasks} & Reconstruction, Link prediction & Reconstruction & Reconstruction \\ 
\textbf{Datasets} & \textit{WordNet}, AstroPh, GrQc & Phylogenetic \cite{hofbauer2016preliminary}, \textit{WordNet}, Diseases & Similar to combinatorial methods \\ 
\textbf{Performance} & High MAP, Low Distortion & Near-perfect on tree-like data & Low Distortion, MAP suboptimal \\ \bottomrule
\end{tabular}
\end{table}
\paragraph{\textbf{{Direct optimization-based methods:}}} These methods focus on learning node embeddings by minimizing a loss function. The main approach is \textit{Poincaré embedding} \cite{nickel2017poincare}, which projects graphs and symbolic data into hyperbolic space using the \textit{Poincaré} ball model. The method learns node embedding by minimizing a task-specific loss function. 
 The update rule for embeddings ensures that embeddings remain inside the \textit{Poincaré} ball. 
%
 For link prediction tasks, the method uses a cross-entropy loss function defined by 
$P((u, v) = 1 \mid \Theta) = \frac{1}{e^{(d(u,v) - r) / t} + 1}$
where \( d(u, v) \) is the hyperbolic distance between node embeddings \( u \) and \( v \), and \( r \) and \( t \) are hyperparameters that adjust the probability distribution.
Evaluated on tasks like link prediction and node classification \cite{nickel2017poincare}, \textit{Poincaré} embedding shows superior performance over \textit{Euclidean} and translational embeddings, particularly in low dimensions. They excel in capturing hierarchical relationships, as seen in datasets like \textit{\textit{WordNet}}, \textit{\textit{\textit{CondMat}}}, and \textit{\textit{HepPh}}. \textit{Poincaré} embedding performs better in terms of mean average precision (MAP) and mean rank scores, demonstrating the advantage of hyperbolic geometry in modeling complex graph structures.

Further work on \textit{Poincaré} embedding was conducted by \cite{nickel2018learning}, who focus on embedding weighted graphs in the  \textit{Lorentz} model. We can also cite order based methods such as  \cite{ganea2018hyperbolic}, which view
hierarchical relations as partial orders defined using a family of nested geodesically convex cones and prove that these entailment cones admit an optimal shape with a closed form expression both in the Euclidean and hyperbolic spaces. To obtain efficient and interpretable algorithms, \cite{law2019lorentzian}  exploit the fact that the squared Lorentzian distance makes it appropriate to represent hierarchies where parent nodes minimize the distances to their descendants.
\cite{suzuki2019hyperbolic} develop Disk Embeddings, which is a framework for embedding DAGs into quasi-metric spaces that outperform existing methods especially in complex DAGs. \cite{NEURIPS2019_82c25591} learn high-precision embedding using an integer-based tiling to represent  points in hyperbolic space with provably bounded numerical error.
of a \textit{Poincar{\'e}} embedding on WordNet Nouns) and learn more accurate embeddings on real-world datasets.
We can cite also  random walk techniques \cite{wang2019hyperbolic}, 
 that adopt architectures that use hyperbolic inner products and hyperbolic distance as proximity measures.

\paragraph{\textbf{{Tree Approximation-Based Methods:}}} These methods embed graphs by transforming them into tree structures before mapping them into hyperbolic space. The main approach is \textit{Combinatorial construction}  \cite{sala2018representation}, which involves embedding a graph into a weighted tree and then mapping this tree into the \textit{Poincaré} ball. 
  This method  was evaluated on several datasets 
 revealing that the embedding dimension scales linearly with the longest path length and logarithmically with the maximum degree in trees. This shows a trade-off between precision and dimensionality, highlighting the method’s suitability for hierarchical structures \cite{sala2018representation}.
 Further advancements in tree approximation-based methods include learning tree structures that approximate the shortest path metric on a given graph, thereby enhancing embedding accuracy 
 \cite{yim2023fitting}.

\paragraph{\textbf{{Multidimensional Scaling-Based Methods:}}} In this category, we can cite mainly    \textit{\textit{H-MDS}} \cite{walter2004h}, which extends multidimensional scaling (\textit{MDS}) \cite{kruskal1978multidimensional} into hyperbolic space, particularly within the  \textit{Lorentz} manifold. \textit{Euclidean} \textit{MDS} methods only recover centered points, but \textit{\textit{H-MDS}} introduces a pseudo-\textit{Euclidean} mean to center points in hyperbolic space. The pseudo-\textit{Euclidean} mean is defined as the local minimum. 
 Once points are centered using this mean, the \textit{\textit{H-MDS}} problem reduces to a standard PCA problem.
\textit{H-MDS} was evaluated \cite{walter2004h} using similar datasets and metrics as the \textit{combinatorial construction}. It demonstrated low distortion, particularly for tree-like datasets, outperforming traditional dimensionality reduction and optimization methods. However, its performance on MAP scores was less optimal due to precision limitations.
Several extensions and enhancement, trying to further minimize  distortion, i.e., the difference between distances in the original graph and distances in the embedding space, are also proposed such as \textit{Lorentzian MDS} \cite{10.1371/journal.pone.0187301}, \textit{\textit{h-MDS}} \cite{pmlr-v80-sala18a}, \textit{hydra} \cite{keller2020hydra}, \textit{HDM} \cite{tabaghi2020hyperbolic}.

For further dimensionality reduction or visualization in one or two dimensions, the authors of 
\cite{huckemann2010intrinsic} suggest using \textit{Principal Geodesics Analysis (PGA)}, a generalization of PCA for non-\textit{Euclidean} manifolds .
To address the quadratic complexity of computing all pairwise distances, the authors of  \cite{shavitt2008hyperbolic} propose methods to select a subset of distances, significantly reducing computational overhead while maintaining accuracy.  In \cite{pmlr-v139-lopez21a} propose to use symmetric spaces to better adapt to dissimilar structures in the graph.
In \cite{gu2018learning}, the authors propose learning embeddings in a product manifold combining multiple copies of model spaces (spherical, hyperbolic, \textit{Euclidean}), providing a space of heterogeneous curvature suitable for a wide variety of structures.

\paragraph{\textbf{{Network model dependent methods:}}}  Inspired by the study of \cite{krioukov2010hyperbolic} on models generating random graphs in hyperbolic spaces, such as the Popularity × Similarity Optimization (PSO) model \cite{papadopoulos2012popularity},
 several works proposed to infer node coordinates in the \textit{Poincaré} disk using Maximum Likelihood Estimation (MLE), maximizing the likelihood that the network was produced by such a model. Several network properties are used for optimizing the obtained embedding.
In \cite{8314557}, the authors rely on common neighbor information.
In \cite{WANG2016609}
, the authors use community structure to compute the angular coordinates.

 In \cite{muscoloni2017machine}, the authors utilize eigenvalue decomposition of the Laplacian matrix to find angular coordinates of nodes in the \textit{Poincaré} disk, while radial coordinates are inferred by employing a network model. Hybrid approaches combining manifold learning and maximum likelihood estimation were explored by  
 \cite{garcia2019mercator}.
 In \cite{papadopoulos2014network}, the authors present \textit{HyperMap} is a simple method for mapping a real network into hyperbolic space based on a recent geometric theory of complex networks.
 It reconstructs the network's geometric growth by estimating the hyperbolic coordinates of new nodes at each step, maximizing the likelihood of the observed network in the model.
\textit{HyperMap-CN} \cite{papadopoulos2015network} is an extension of \textit{HyperMap} that also uses the number of common neighbors between the nodes.

\noindent \underline{\textbf{{{2-Deep learning methods}}}}
Graph embedding methods based on deep learning have demonstrated remarkable progress in capturing complex graph structures. The integration of hyperbolic geometry into deep learning for graph data has significantly advanced, particularly since the work of \cite{ganea2018hnn}. Hyperbolic geometry offers an effective framework for encoding hierarchical and tree-like structures, enabling deep learning models to represent graph data with greater fidelity. Below, we discuss methods grouped into \textit{GNN}-based approaches, graph autoencoders, and spatio-temporal models.

\paragraph{\textbf{{GNN-based methods:}}} In this category, we have \textit{Hyperbolic Graph Neural Networks (HGNNs)} \cite{liu2019hyperbolic} that extend traditional graph neural networks (GNNs) to \textit{Riemannian} manifolds, specifically the \textit{ \textit{Lorentz}} and \textit{Poincaré ball} models. Building on the framework of the \textit{Euclidean} \textit{Vanilla GCN} \cite{kipf2017semi}, \textit{HGNNs} utilize differentiable exponential and logarithmic maps to perform message passing. At each layer \(k\), the node representation \(h_v^k\) is mapped to the tangent space of a reference point \(x'\) using the logarithmic map \(\log_{x'}\), aggregated, and then mapped back to the manifold via the exponential map \(\exp_{x'}\). The updated representation \(h_u^{k+1}\) is given by:

\[
h_u^{k+1} = \sigma\left(\exp_{x'} \left(\sum_{v \in I(u)} \tilde{A}_{uv} W_k \log_{x'} (h_v^k)\right)\right),
\]

where \(W_k\) are trainable parameters, \(\tilde{A}\) is the normalized adjacency matrix, and \(I(u)\) denotes the in-neighbors of node \(u\). This framework allows \textit{HGNN} to perform message passing in the tangent space and map results back to hyperbolic space effectively.

\textit{HGNNs} operate in both \textit{ \textit{Lorentz}} and \textit{Poincaré} models, with curvature fixed to \(-1\). In the \textit{Poincaré} model, pointwise non-linearities are used for activation, whereas in the \textit{ \textit{Lorentz}} model, points are projected to the \textit{Poincaré} space for activation and then mapped back. The output consists of hyperbolic node embeddings used for node-level and graph-level predictions, leveraging cross-entropy and mean squared error loss functions, respectively. This approach demonstrate superior performance in preserving graph structures compared to \textit{Euclidean} methods, excelling in graph classification tasks. Notably, the  \textit{Lorentz} model outperformed the \textit{Poincaré} model, especially in higher-dimensional settings and datasets such as \textit{Zinc} for molecular property prediction and \textit{Ether/USDT} for time series. \textit{HGNNs} showcase robust performance in complex scenarios, advancing the application of hyperbolic geometry in GNNs \cite{liu2019hyperbolic}.

\textit{Hyperbolic Graph Convolutional Neural Networks (HGCNs)}  \cite{chami2019hyperbolic}, extend Graph Convolutional Networks (\textit{GCNs}) to hyperbolic geometry, achieving significant advancements in preserving hierarchical structures and tree-like data. \textit{HGCNs} operate in the hyperboloid model with trainable curvature, where each layer $\ell$ functions at a different curvature $-1/K_\ell$ ($K_\ell > 0$) centered at $o = \{\sqrt{K_\ell}, 0, \ldots, 0\}$. \textit{Euclidean} input features $x_{i0,E}$ are mapped to hyperbolic space using the exponential map $\exp_{o K_\ell}$. Each layer performs a transformation of node features in the tangent space using the formula:
\[
h_{i,H}^\ell = \left( W_\ell \otimes_{K_\ell^{-1}} x_{i,H}^{\ell-1} \right) \oplus_{K_\ell^{-1}} b_\ell,
\]
where $W_\ell$ and $b_\ell$ are the weights and bias parameters of layer $\ell$. \textit{HGCNs} employ a hyperbolic attention-based mechanism to aggregate features of neighboring nodes:
\[
x_{i,H}^\ell = \sigma_{\otimes_{K_\ell^{-1}}, K_\ell} \left( \exp_{K_\ell} \left( \sum_{j \in N(i)} w_{ij} \log_{K_\ell} x_{j,H} \right) \right),
\]
where $N(i)$ represents the neighbors of node $i$, and $w_{ij}$ are attention weights computed using an \textit{Euclidean} multi-layer perceptron (MLP) similar to the \textit{Graph Attention Network (GAT)} \cite{velivckovic2017graph}. Unlike fixed-curvature models, \textit{HGCNs} assign distinct curvature values $-1/K_\ell$ to each layer, enhancing flexibility and alignment with the dataset’s geometry. For link prediction tasks, \textit{HGCNs} employ the \textit{Fermi-Dirac} decoder to compute edge probability scores, while for node classification, the output embeddings are projected to the tangent space for \textit{Euclidean} multinomial logistic regression. \textit{HGCNs} were evaluated on datasets such as \textit{CoRA} \cite{sen2008collective}, \textit{PubMed} \cite{namata2012query}, \textit{SIR} \cite{anderson1991infectious}, \textit{PPI} \cite{szklarczyk2016string}, and \textit{Airport} (a flight network), where they outperform baseline methods such as \textit{GCN}, \textit{GAT}, and \textit{Poincaré} embeddings, especially on datasets with low $\delta$-hyperbolicity \cite{chami2019hyperbolic}. Its attention mechanisms and trainable curvature demonstrated improved preservation of hierarchies, class separability, and tree structures. In particular, \textit{HGCNs} achieve state-of-the-art results in link prediction and node classification tasks, showcasing robust performance across varying data geometries \cite{chami2019hyperbolic}.


\textit{Hyperbolic-to-Hyperbolic Graph Convolutional Networks (H2H-GCNs)} \cite{9578363}, address the limitations of hyperbolic GCNs that rely on tangent space approximations for feature transformation and neighborhood aggregation by performing all operations directly within hyperbolic manifolds. The model projects \textit{Euclidean} input features into the  \textit{Lorentz} manifold using the exponential map and applies  \textit{Lorentz} linear transformations at each layer $\ell$, defined as:
\[
\overline{h}_i^{\ell} = W_\ell h_i^{\ell-1}, \quad \text{where} \quad W_\ell = \begin{pmatrix} 1 & 0^\top \\ 0 & \mathbf{W}_\ell \end{pmatrix}, \quad \mathbf{W}_\ell^\top \mathbf{W}_\ell = I.
\]
Neighborhood aggregation is performed using the Einstein midpoint in the Klein model, followed by activation in the \textit{Poincaré} model. By avoiding tangent space approximations, \textit{H2H-GCNs} reduce distortion and improves performance, particularly in low-dimensional hyperbolic spaces.
\textit{H2H-GCNs} \cite{9578363} demonstrate superior results across datasets like \textit{Disease}, \textit{Cora}, \textit{PubMed}, and \textit{Airport} for tasks such as link prediction and node classification. It achieved top AUC scores for link prediction and high F1 scores for node classification, outperforming models like \textit{HGCNs} and other baselines. For graph classification on synthetic and molecular graphs, \textit{H2H-GCNs} excel in preserving hyperbolic structures, showcasing advantages over tangent space-based models like \textit{HGNNs}, which suffer from distortion in higher dimensions. Its robust performance in both low- and high-dimensional settings establishes \textit{H2H-GCN} as a significant advancement in hyperbolic graph learning.

Several \textit{GNN} variants operate directly on the hyperbolic manifold, leveraging non-\textit{Euclidean} geometry to better capture hierarchical and complex graph structures. For instance, \textit{k-GCN} \cite{pmlr-v119-bachmann20a} generalizes GCNs to constant curvature spaces using gyro-barycentric coordinates, enabling smooth interpolation between \textit{Euclidean} and non-\textit{Euclidean} geometries. Similarly, \textit{HAT} \cite{zhang2021hyperbolic} introduces hyperbolic attention-based aggregation, computing attention coefficients based on hyperbolic distances. \textit{GIL} \cite{NEURIPS2020_551fdbb8} embeds graphs in both \textit{Euclidean} and hyperbolic spaces for flexible geometry-aware representations, while \textit{Q-GCN} \cite{NEURIPS2022_16c628ab} extends GCNs to pseudo-Riemannian manifolds to handle mixed topologies. Adaptive approaches like \textit{ACE-HGCNN} \cite{9679192} use reinforcement learning to optimize curvature, and \textit{NHGCN} \cite{fan2022nested} employs nested hyperbolic spaces for efficient representation learning. Additionally, \textit{$\kappa$HGCN} \cite{kHGCN} models tree-likeness by learning discrete and continuous curvature.

Beyond hyperbolic geometry, \textit{FMGNN} \cite{deng2023fmgnnfusedmanifoldgraph} fuses embeddings from multiple Riemannian manifolds, aligning different geometric spaces for enhanced learning. Finally, \textit{LGCN} \cite{zhang2021lorentzian} redefines graph operations using \textit{Lorentzian} distance, reducing distortion and preserving hierarchical structures.

\paragraph{\textbf{{Graph autoencoder-based methods:}}} Graph autoencoder-based methods have been successfully extended to hyperbolic spaces, providing a novel approach for learning unsupervised representations of graph data. In this category, we can cite \textit{Poincaré Variationnal Auto-Encoders ($\mathcal{P}$-VAE)} \cite{mathieu2019continuous}, which extends Variational Autoencoders (VAEs) to hierarchical or tree-structured data by mapping it into a hyperbolic latent space, specifically the \textit{Poincaré} ball $B_d^c$ with constant negative curvature $-c$. VAEs, introduced by \cite{kingma2013auto}, are generative models that encode input data into a latent space, typically modeled as a Gaussian distribution in \textit{Euclidean} space, and decode it back to the original space. The key idea is to optimize the Evidence Lower Bound (ELBO), balancing reconstruction accuracy with latent space regularization, allowing for the generation of new, similar data points.

In $\mathcal{P}$-VAE, the latent space is modeled using Riemannian normal and wrapped normal distributions. The encoder outputs the Fréchet mean $\mu$ in the \textit{Poincaré} ball and standard deviation $\sigma$. To maintain hyperbolicity, the encoder employs an exponential map layer, while the decoder utilizes a gyroplane layer, which involves hyperbolic affine transformations that map the latent representation back to \textit{Euclidean} space. $\mathcal{P}$-VAE also adapts the reparameterization trick and ELBO for hyperbolic geometry, allowing for better preservation of hierarchical structures inherent in the data. It showed improved performance on graph datasets, particularly in link prediction tasks, effectively capturing the hierarchical relationships within the data compared to \textit{Euclidean} methods \cite{mathieu2019continuous}.

\textit{Hyperbolic Graph Convolutional Autoencoder (HGCAE)} \cite{park2021unsupervised} is a hyperbolic version of graph autoencoders that operates on both the \textit{Poincaré} ball and  \textit{Lorentz} models, with trainable curvature across layers. The architecture is similar to \textit{HGCN} with attention-based aggregation. For node $i$ at layer $\ell$, the message passing operation is:

\[
z_i^\ell = \exp_{K_\ell}^{o} \left( \sum_{j \in N(i)} \alpha_{ij}^\ell \left( W_\ell \log_{K_\ell}^o (h_j^\ell) + b_\ell \right) \right),
\]

where $h_j^\ell$ is the embedding of node $j$ from the previous layer, $K_\ell$ is the curvature at layer $\ell$, $W_\ell$ and $b_\ell$ are the weight matrix and bias, and $\alpha_{ij}^\ell$ represents the attention score. The node representation $h_i^{\ell+1}$ is obtained by applying an activation function, such as ReLU, in the \textit{Poincaré} ball.

The encoder reconstructs the adjacency matrix $A$, while the decoder reconstructs the \textit{Euclidean} node attributes $X_\text{Euc}$. The total loss function combines cross-entropy loss for the adjacency matrix and mean squared error for node attributes
$L = L_{\text{REC}-A} + \lambda L_{\text{REC}-X},$
 where $\lambda$ controls the balance between structural and attribute reconstruction.

\textit{HGCAE} was evaluated \cite{park2021unsupervised}  on link prediction and node clustering tasks using datasets like \textit{CoRa}, \textit{Citeseer}, \textit{Wiki}, \textit{PubMed}, and \textit{BlocCatalog}, outperforming state-of-the-art models like \textit{GAE}, \textit{VGAE}, \textit{ARGA}, and \textit{DBGAN}. The \textit{Poincaré} (\textit{HGCAE-P}) and hyperboloid (\textit{HGCAE-H}) variants achieved superior AUC and average precision for link prediction and higher accuracy and NMI for node clustering. \textit{HGCAE-H} generally performed best, with some exceptions on \textit{Wiki} and \textit{PubMed}. Visualizations on \textit{CoRa} demonstrated effective node clustering near the hyperbolic space boundary, showcasing the model’s ability to capture complex graph structures.

Extending beyond these architectures, additional methods explore diverse autoencoder-based approaches in hyperbolic spaces.	\textit{GCM-AAE} \cite{GRATTAROLA2019105511} introduces an adversarial autoencoder designed for constant-curvature \textit{Riemannian} manifolds (CCMs), effectively embedding hierarchical and circular structures. By aligning the aggregated posterior with a probability distribution on a CCM, the model enhances representation learning across varying curvature spaces.

In another direction, \textit{PWA} \cite{ovinnikov2019poincar} reformulates Wasserstein autoencoders within the \textit{Poincar\'e ball}, leveraging hyperbolic latent spaces to impose a hierarchical structure on learned representations. The approach demonstrates strong results in graph link prediction by effectively structuring latent spaces according to data hierarchy.

Finally, \textit{Hyperbolic NF} \cite{bose2020latent} introduces normalizing flows into hyperbolic spaces, overcoming limitations of \textit{Euclidean}-based flows in stochastic variational inference. By employing coupling transforms on the tangent bundle and Wrapped Hyperboloid Coupling (WHC), Hyperbolic NF achieves expressive posteriors, outperforming \textit{Euclidean} and hyperbolic VAEs on hierarchical data density estimation and graph reconstruction.

\paragraph{\textbf{{Spatio-temporal GNN-based methods:}}} In this category, we have \textit{Hyperbolic Temporal Graph Network (HTGN)} \cite{DBLP:journals/corr/abs-2107-03767}, which is  designed for discrete-time graphs represented as snapshots from a temporal graph \( G \). Each snapshot \( G_t = (V_t, A_t) \) includes the current node set \( V_t \) and its adjacency matrix \( A_t \). Operating in the \textit{Poincaré} ball model with learnable curvature \( c \), \textit{HTGN} consists of three components: (1) a Hyperbolic Graph Neural Network (\textit{HGNN}) for topological dependency extraction, (2) Hyperbolic Temporal Attention (\textit{HTA}) to aggregate historical information, and (3) a Hyperbolic Gated Recurrent Unit (\textit{HGRU}) for capturing sequential patterns. \textit{HTGN} introduces a Hyperbolic Temporal Consistency (\textit{HTC}) constraint to ensure smooth embedding transitions over time. 

\begin{sidewaystable}[!h]
\centering
\caption{Summary of the main Deep Learning Hyperbolic Graph Embedding Methods.}
\label{tab:approaches_manifolds_descriptions_tasks_datasets}
\footnotesize
\begin{tabular}{m{2.2cm}p{1.9cm}p{2cm}p{3cm}p{3cm}p{3cm}p{3cm}}
\toprule
\textbf{Category} & \textbf{Approach} & \textbf{Manifold} & \textbf{Objective} & \textbf{Tasks} & \textbf{Datasets} & \textbf{Performance} \\ \midrule

\multirow{3}{*}{\centering \textbf{GNN}}
& \textit{HGNN} \cite{liu2019hyperbolic}
& \textit{Poincaré}, \textit{Lorentz}
& Learn node representations in hyperbolic space
& Node regression, Graph regression, Node classification, Graph classification
& \textit{Zinc}, \textit{Ether USDT}
& Outperformed Euclidean methods  \\ \cline{2-7}

&\textit{HGCN} \cite{chami2019hyperbolic}
& Hyperboloid
& Extend GCNs to hyperbolic geometry
& Link prediction, Node classification
& \textit{Cora}, \textit{PubMed}, \textit{SIR}, \textit{PPI}, \textit{Airport}
&low $\delta$-hyperbolicity datasets  \\ \cline{2-7}

& \textit{H2H-GCN} \cite{9578363}
& \textit{Lorentz} Manifold
& Preserve hyperbolic structures
& Link prediction, Node classification, Graph classification
& \textit{Disease}, \textit{Cora}, \textit{PubMed}, \textit{Airport}
& High AUC and F1 scores  \\ \midrule

\multirow{2}{*}{\centering   \textbf{Autoencoder} }
& \textit{$\mathcal{P}$-VAE} \cite{mathieu2019continuous}
& \textit{Poincaré} Ball
& Capture hierarchical data structure
& Link prediction, Reconstruction
& Various graph datasets
& Improved performance in link prediction  \\ \cline{2-7}

& \textit{HGCAE} \cite{park2021unsupervised}
& \textit{Poincaré}, \textit{Lorentz}
& Reconstruct adjacency and attributes
& Node clustering, Link prediction
& \textit{Cora}, \textit{Citeseer}, \textit{Wiki}, \textit{PubMed}, \textit{BlocCatalog}
& Superior AUC and average precision  \\ \midrule

\multirow{2}{*}{\shortstack{\textbf{Spatio-}\\ \textbf{Temporal}\\ \textbf{GNN}}}
& \textit{HTGN} \cite{DBLP:journals/corr/abs-2107-03767}
& \textit{Poincaré} Ball
& Capture temporal patterns in graphs
& Link prediction, New link prediction
& \textit{HepPh}, \textit{COLAB}, \textit{FB}
& Robust performance in sparse graphs  \\ \cline{2-7}

& \textit{HGWaveNet} \cite{Bai_2023}
& \textit{Poincaré}, \textit{Lorentz}
& Learn representations from dynamic graphs
& Link prediction, New link prediction
& \textit{Enron}, \textit{DBLP}, \textit{MovieLens}
& Outperformed static and dynamic models\\ \bottomrule
\end{tabular}
\end{sidewaystable}

\FloatBarrier
\textit{HTGN} minimizes an overall loss function combining hyperbolic distance for temporal smoothness and a homophily loss to preserve graph structure. Evaluated on datasets like \textit{HepPh}, \textit{COLAB}, and \textit{FB}, \textit{HTGN} outperformed several \textit{Euclidean} baselines in link prediction and particularly excelled in new link prediction tasks, demonstrating robust inductive learning even on sparse graphs.

\textit{HGWaveNet} \cite{Bai_2023} is a hyperbolic graph neural network framework designed for discrete dynamic graphs, inspired by \textit{WaveNet} \cite{DBLP:journals/corr/OordDZSVGKSK16} and Graph \textit{WaveNet} \cite{wu2019graph}. It operates in the \textit{Poincaré} ball or  \textit{Lorentz} manifolds with learnable curvature and consists of several key modules: (1) an Hyperbolic Diffusion Graph Convolution (\textit{HDGC}) module that learns node representations from both direct and indirect neighbors in a snapshot through a diffusion process, performing \( K \) diffusion steps to consider longer paths in message propagation; (2) An Hyperbolic Dilated Causal Convolution (\textit{HDCC}) module that aggregates historical information while respecting causality, allowing each state to depend only on past and present states. This module uses a fixed kernel size to update node representations across snapshots, effectively capturing long-range dependencies; (3) an Hyperbolic Gated Recurrent Unit (\textit{HGRU}): Similar to the one in \textit{HTGN}, it processes the current node representations and historical hidden states; and (4) an Hyperbolic Temporal Consistency (\textit{HTC}) module that Ensures stability in the embeddings over time. \textit{HGWaveNet} was evaluated on link prediction tasks using datasets such as \textit{Enron}  and \textit{DBLP}
outperforming multiple static and dynamic models. 
 Building on the previously discussed spatio-temporal GNN-based methods, several other approaches leverage hyperbolic geometry for improved representation learning over dynamic graphs.   \textit{ST-GCN} \cite{10.1145/3394171.3413910} extends traditional spatio-temporal GCNs by integrating \textit{Poincaré} geometry, effectively modeling latent anatomical structures in human action recognition while optimizing projection dimensions in the Riemann space. By leveraging hyperbolic embeddings, \textit{ST-GCN} enhances structural representation efficiency while reducing model complexity. In another direction, \textit{HVGNN} \cite{sun2021hyperbolic} introduces a hyperbolic variational framework to model both dynamics and uncertainty in evolving graphs. The method combines a Temporal GNN (\textit{TGNN}) with a hyperbolic graph variational autoencoder to generate stochastic node representations, ensuring robust performance across dynamic settings. 
  By leveraging hyperbolic graph attention networks, it effectively captures spatial and temporal dependencies among stocks, significantly enhancing profitability and risk-adjusted returns through structured hyperbolic representations. Finally, \textit{DHGAT} \cite{LI2024127038} proposes a novel spatiotemporal self-attention mechanism based on hyperbolic distance that 
  achieves superior performance in multi-step link prediction tasks, particularly excelling at identifying new and evolving links within dynamic graphs.

\section{An Anomaly Detection Framework for Hyperbolic Embeddings}
\label{section:solution}
\begin{figure}[t]
    \centering
    \includegraphics[width=0.9\textwidth]{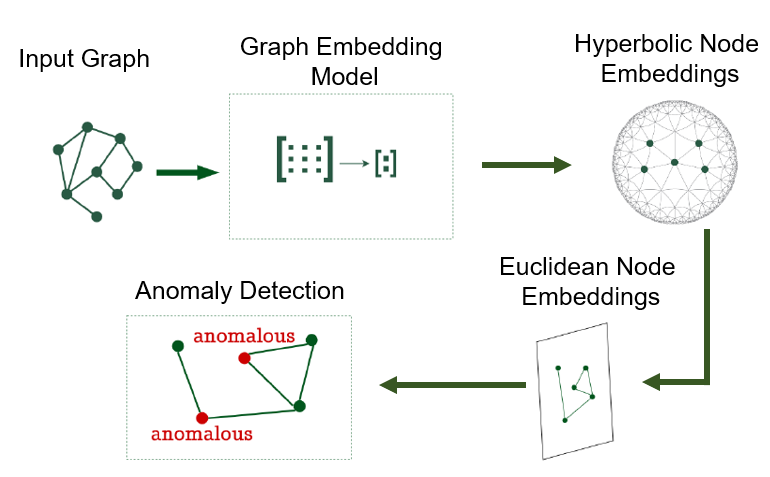}
    \caption{Global architecture of the anomaly detection framework.}
    \label{fig:graph}
\end{figure}

Embedding graphs into hyperbolic spaces is gaining attention because it helps capture the hierarchical and scale-free nature of many real-world networks. These networks often have a few highly connected nodes and many less connected ones, and they tend to form tightly-knit groups. Hyperbolic geometry is well-suited for representing these properties because it can model hierarchical structures with low distortion 
\cite{krioukov2010hyperbolic}. Despite these advantages, embedding graphs in hyperbolic spaces introduces several challenges. Unlike \textit{Euclidean} spaces, hyperbolic spaces do not possess vector space properties, making basic operations such as vector addition, matrix-vector multiplication, and gradient-based optimization problematic \cite{zhang2021lorentzian}. These operations are crucial for machine learning algorithms and may result in outputs that lie outside the manifold, complicating the embedding process \cite{peng2021hyperbolic}. Addressing these challenges requires innovative methods tailored to the unique characteristics of hyperbolic geometry.  In this section, we propose a unified framework for node-level anomaly detection in homogeneous static graphs, enabling the testing and comparison of various embedding methods alongside different anomaly detection techniques.
The overall architecture of our framework is illustrated in Figure \ref{fig:graph}. This architecture is divided into two primary components:
\begin{itemize}
       \item \textbf{Graph Embedding Model:} 
    This model generates node embeddings in a hyperbolic space (e.g., \textit{Poincaré} ball,  \textit{Lorentz} model), which are then mapped to the \textit{Euclidean} space. This setup allows flexibility in evaluating various state-of-the-art hyperbolic embedding techniques.
    \item \textbf{Anomaly Detection Mechanism:} The \textit{Euclidean} embeddings are used by an anomaly detection algorithm (supervised or unsupervised) to classify nodes as benign (normal) or malicious(abnormal).
\end{itemize}

In this framework, we evaluated six models: \textit{H2H-GCN} \cite{9578363}, \textit{HGCAE} \cite{park2021unsupervised}, \textit{HGCN} \cite{chami2019hyperbolic}, \textit{HGNN} \cite{liu2019hyperbolic}, \textit{Poincaré} \cite{nickel2017poincare}, and $\mathcal{P}$-VAE \cite{mathieu2019continuous}.
In order to ensure reproducibility and facilitate comparative analysis, we summarize the hyperparameters used for each model in Table~\ref{table:hyperparameters}. Key hyperparameters such as learning rate, weight decay, and dropout were fixed, with adjustments based on the dataset and embedding dimensionality.
For anomaly detection, we applied a two-phase approach with models like \textit{$\mathcal{P}$-VAE} and \textit{HGCAE}. The first phase involves training the models, while the second phase applies various anomaly detection algorithms (e.g., Random Forest and k-Nearest Neighbors) to classify nodes as benign or anomalous. To support these efforts, we developed \textbf{\textit{Ghypeddings}}, a Python library designed to embed static graph nodes into hyperbolic space. Short for Graph Hyperbolic Embeddings, \textit{Ghypeddings} consolidates all six hyperbolic graph models and integrates some anomaly detection algorithms. Built on the PyTorch framework, this library provides a comprehensive solution for hyperbolic graph embeddings and is publicly available. \textit{Ghypeddings} streamlines the process of building and loading graphs, training models, and saving/loading them for future use, thus improving the usability and reproducibility of the models.

\begin{table}[t]
\centering
\caption{Hyperparameters for each model used in the study.}
\label{table:hyperparameters}
\footnotesize
\begin{tabular}{l p{5cm}p{2.5cm}p{2cm}}
\toprule
\textbf{Model} & \textbf{Layers and Attention Mechanism} & \textbf{Initialization} & \textbf{Optimizer} \\ \midrule
\textit{H2H-GCN} & Two layers with SELU activation. & Uniform Bias, Orthogonal Weights & Riemannian SGD \\ 
\textit{HGCAE} & Two-layer encoder and decoder with ReLU activation, sparse attention. & - & Riemannian Adam \\ 
\textit{HGCN} & Two layers with ReLU activation and dense attention. & - & Riemannian Adam \\ 
\textit{HGNN} & Two layers with LeakyReLU activation. & Xavier Initialization & Riemannian AMSGrad \\ 
Poincaré & Optimized similarly to HGCAE. & - & Riemannian Adam \\
\textit{$\mathcal{P}$-VAE} & Two GCN layers and one Möbius layer in the encoder; one gyroplane and two FFNN layers in the decoder. & Wrapped Normal & - \\ \bottomrule
\end{tabular}
\end{table}

\section{Experiments}
This section presents a comprehensive evaluation of anomaly detection models using multiple datasets. It begins by describing the datasets used, followed by a discussion of the \textit{Euclidean} embedding models included for comparison to benchmark the performance of hyperbolic embeddings. The preprocessing steps applied to the datasets are detailed to ensure reproducibility. Key evaluation metrics, such as accuracy and F1-score, are introduced to provide a consistent basis for assessing model performance. Finally, the results analyze and compare the performance of hyperbolic and \textit{Euclidean} embedding models, highlighting their strengths, trade-offs, and computational efficiency.
\label{section:expermient}

\noindent \textbf{\underline{Datasets:}}
For evaluating anomaly detection models across various domains, we selected a diverse set of datasets to ensure robustness and consistency. The datasets used are listed in table \ref{tab:datasets}. For each dataset, data is partitioned into training (70\%), validation (20\%), and testing (10\%) sets.

\noindent \textbf{\underline{Metrics:}} The effectiveness of the models on anomaly detection are assessed using:
 (1) the classification Outcomes, 
 based on  the commonly used metrics summarized in Table \ref{tab:metrics}, and  
the computational efficiency, 
particularly regarding training time (TT), which represents the duration in seconds that the model requires to learn from the training dataset, including the optimization of parameters for both graph embedding and anomaly detection. 

\begin{table}[t!]
\centering\caption{Evaluation Metrics for Anomaly Detection.}
    \label{tab:metrics}
    \footnotesize
        \begin{tabular}{p{1cm}p{7cm}p{2cm}}
    \toprule
    \textbf{Metric} & \textbf{Description} & \textbf{Formula} \\ \midrule
    Accuracy & Measures the overall correctness of the model's predictions. &
    \(\frac{TP + TN}{TP + TN + FP + FN}\) \\

    Precision & Indicates the accuracy when the model predicts an anomaly. &
    \(\frac{TP}{TP + FP}\) \\
    Recall & Measures the model's ability to detect all actual anomalies. &
    \(\frac{TP}{TP + FN}\) \\
    F1 Score & The harmonic mean of precision and recall, balancing the two. &
    \(\frac{2 \times \text{Precision} \times \text{Recall}}{\text{Precision} + \text{Recall}}\) \\
    AUC & Evaluates the model's performance by plotting the true positive rate against the false positive rate at various thresholds. &
    - \\ \bottomrule
       \end{tabular}\\
        \scriptsize
    TP: True positives. TF: True negatives. FP: False positives. FF: False negatives.
  \end{table}

\noindent \textbf{\underline{Euclidean Baselines:}} 
To evaluate the effectiveness of hyperbolic embeddings, we compare them with two prominent \textit{Euclidean} embedding models:
  \textbf{GraphSage} \cite{hamilton2017inductive}, which  has demonstrated strong performance in graph representation tasks, and  
 \textbf{DOMINANT} \cite{ding2019deep}, which integrates Graph Convolutional Networks (GCNs) with Autoencoders to detect outlier nodes in graphs. \textit{DOMINANT} employs a GCN-based encoder and two GCN-based decoders: one reconstructs node features, and the other reconstructs the adjacency matrix. The model outputs an outlier score that combines reconstruction errors from both decoders.
 We selected these models due to their strong performance on node anomaly detection tasks and their compatibility with diverse datasets.

\begin{table}[t]
    \centering
    \caption{Datasets}
    \label{tab:datasets}
    \footnotesize
    \begin{tabular}{l p{4.5cm}p{4.5cm}}
    \toprule
    \textbf{Dataset} & \textbf{Description} & \textbf{Key Details} \\ \midrule
        \textbf{Darknet} & network traffic with malicious activities &
    ~30GB size, millions of flows, 20+ features, 85\% benign, 15\% malicious \\
        \textbf{CICDDoS2019} & Network traffic with malicious activities &
    80 features, 12 different types of malicious activities\\         \textbf{DGraphFin} & Financial transaction graph for fraud detection. &
    500K nodes, 1.5M edges, ~1\% fraud. \\
        \textbf{Elliptic} & Bitcoin transactions containing illicit activities &
    203,769 transactions, 234,355 edges, 2\% illicit, 166 features \\
        \textbf{Cora} & Citation network of research papers &  2,708 nodes, 5,429 edges, 7 classes,  1,433 features \\
    \textbf{YelpNYC} & Yelp reviews for detecting fake reviews &
    359K reviews, includes user, business, and review data, fraud labels. \\ \bottomrule

    \end{tabular}

\end{table}

\noindent \textbf{\underline{Results and Discussion:}}
The results of all the algorithms are depicted in Table \ref{tab:darknet} for the \textit{Darknet} datatset, Table \ref{tab:cicddos} for \textit{CICDDoS2019} dataset, Table \ref{tab:elliptic} for
\textit{Elliptic} datatset, Table \ref{tab:yelpnyc} \textit{YelpNYC} datatset, Table\ref{tab:cora} for  \textit{Cora} datatset, and Table \ref{tab:dgraphfin}  for \textit{DGraphFin} dataset.

In addition to the comparison, we are working on selecting the classification optimal algorithms for autoencoder-based embedding models, specifically \textit{HGCAE} and \textit{$\mathcal{P}$-VAE}, to maximize their performance.
In all tables, the best results for each metric are marked in bold and underlined, while the second-best results are underlined.\\
Across all datasets, we kept the embedding dimension fixed at 10, as this configuration provided consistently better results across most models.

\textbf{Performance of Autoencoder-based models with different classifiers:} We used these models with
 Random Forest (RF), k-Nearest Neighbors (kNN),
Agglomerative Clustering (AC), and
 Gaussian Mixture (GM) classifiers.
According to Tables \ref{tab:darknet}- \ref{tab:dgraphfin}, 
 \textit{HGCAE}, especially when integrated with kNN and GM, generally demonstrates solid performance, achieving relatively high accuracy and F1-scores across most datasets, but it faces challenges in training time (TT) efficiency.  For instance, on the \textit{Cora dataset}, \textit{HGCAE}+KNN achieves an accuracy of 78\% and an F1-score of nearly 80\%, but requires a training time of 8 seconds, highlighting its computational demands. This trend persists across other datasets, where \textit{HGCAE} variants show comparable performance but necessitate significantly longer training periods, notably on more complex datasets like \textit{Darknet}, where TT reaches 23 seconds. In contrast, \textit{$\mathcal{P}$-VAE}, particularly when paired with Random Forest(RF) or Agglomerative Clustering(AC) classifiers, exhibits excellent performance metrics, especially in terms of accuracy and AUC scores, indicating strong anomaly detection capabilities. For instance, \textit{$\mathcal{P}$-VAE}+RF achieves the highest accuracy on datasets such as \textit{YelpNYC} 88\%  and \textit{Cora} 86\%. However, this performance comes at the cost of even longer training times, with values such as 32 seconds on Cora and nearly 34 seconds on \textit{YelpNYC}, due to the model's complexity and the additional computation required for variational inference. This underscores the computational intensity of \textit{$\mathcal{P}$-VAE} approaches in anomaly detection tasks, where the benefits of accuracy and robustness must be balanced against their high computational overhead.  Highlighting \textit{$\mathcal{P}$-VAE}’s high F1-scores on datasets like \textit{Darknet} and \textit{Elliptic} is essential, as these scores underscore its robustness in complex anomaly detection tasks. For instance, \textit{$\mathcal{P}$-VAE} combined with AC achieves an impressive F1-score of 92\% on the \textit{Darknet} dataset, indicating that it is particularly effective at identifying anomalies in high-dimensional, complex network data.
 Similarly, on the  \textit{Elliptic} dataset, \textit{$\mathcal{P}$-VAE}+AC achieves an F1-score of 94\%, one of the highest across all tested models, reflecting its robustness in detecting fraudulent transactions in financial networks.

\textbf{Comparison of the Hyperbolic Embedding Models:} %
Tables \ref{tab:darknet}- \ref{tab:dgraphfin} show that the \textit{Poincaré} model consistently achieves low training times (TT), especially notable on datasets such as \textit{Darknet} and \textit{Elliptic}, where it maintains a training time below 2.5 seconds. While its recall scores are generally high indicating a strong ability to identify anomalies , the model shows relatively lower precision on most datasets. For instance, on the \textit{Elliptic} dataset, \textit{Poincaré} achieves a recall of 95\% but has a lower precision of 59\%, which could imply a higher false-positive rate.
\textit{HGNN} and \textit{HGCN} models demonstrate impressive performance metrics across both the \textit{Darknet} and \textit{CICDDoS2019} datasets. \textit{HGNN}, for instance, attains high F1-scores and recall rates, reaching 88.92\% F1-score on \textit{Darknet} and 93\% on \textit{CICDDoS2019}. These results highlight \textit{HGNN}'s effectiveness in identifying and classifying anomalies accurately, although at a higher computational cost (e.g., 19.96 seconds on \textit{CICDDoS2019}). \textit{HGCN} generally achieves high precision scores, especially on datasets like \textit{CICDDoS2019} (94.95\%), indicating its robustness in reducing false positives.
\textit{H2H-GCN} stands out on the \textit{Elliptic} dataset with an accuracy of 89.71\% and precision of 96.55\%, reflecting strong performance in detecting financial anomalies, a critical feature in applications such as fraud detection. This high performance on Elliptic, combined with moderate training times, suggests \textit{H2H-GCN}'s applicability to complex network structures with limited computational trade-offs.

\textbf{Performance of the \textit{Euclidean} Embedding Models:} \textit{Euclidean} embedding methods typically underperformed compared to hyperbolic embeddings, especially on datasets with hierarchical or highly structured data, where the limitations of \textit{Euclidean} space in capturing latent hierarchies become apparent. However, some models, particularly \textit{GraphSage}, occasionally achieved competitive results, demonstrating relatively high precision and low training time on less structured datasets. Overall, the performance of \textit{Euclidean} embedding models varied significantly by dataset.
    \textit{DOMINANT}, 
    both in 10\% and 50\% variants, exhibited high recall across several datasets but suffered in accuracy and precision, leading to an inflated number of false positives. For example, on the \textit{Darknet} dataset, \textit{DOMINANT} (10\%) achieved a recall of 93\%, indicating strong anomaly detection capability. However, its accuracy was only 53.68\%, underscoring the model’s tendency to misclassify normal points as anomalies. A similar trend was observed across other datasets, where \textit{DOMINANT} consistently produced high recall scores at the expense of precision and accuracy, highlighting its capacity to capture anomalies broadly but with limited specificity.
     \textit{GraphSage}, in contrast,
     demonstrated a better balance between precision, recall, and overall accuracy. On simpler datasets such as Cora and \textit{YelpNYC}, \textit{GraphSage} achieved competitive scores across evaluation metrics. For instance, in the \textit{Cora} dataset, \textit{GraphSage} obtained an accuracy of 88\% and an F1-score of 87.6\%, indicating that it effectively captured anomalies while maintaining a lower rate of false positives. On the \textit{YelpNYC} dataset, where it achieved a precision of 93\%, the model demonstrated an ability to correctly identify true anomalies with minimal misclassifications. The balanced performance across multiple metrics suggests that \textit{GraphSage} may offer robust anomaly detection capabilities within \textit{Euclidean} spaces for datasets without strong hierarchical patterns.

In terms of computational efficiency, \textit{Euclidean} embeddings offer significant advantages. Both \textit{DOMINANT} and \textit{GraphSage} exhibit notably low training times, rendering them suitable for large-scale applications or situations where computational resources are limited. For example, \textit{DOMINANT} required less than 1.5 seconds of training on several datasets, whereas \textit{GraphSage} consistently showed shorter training times compared to hyperbolic models, such as \textit{$\mathcal{P}$-VAE} and \textit{H2H-GCN}, which can exceed 30 seconds in training time. This efficiency makes \textit{Euclidean} models particularly useful in scenarios where quick model deployment and lower computational overhead are priorities.

\begin{table}
  \centering
  \caption{Performance on the \textit{Darknet} Dataset.}
  \label{tab:darknet}
  \footnotesize
  \renewcommand{\arraystretch}{1.1}
  \begin{tabular}{clcccccc}
    \cmidrule[\heavyrulewidth](l){2-8}
    & \textbf{Model} & \textbf{Accuracy} & \textbf{F1-score} & \textbf{Recall} & \textbf{Precision} & \textbf{AUC} & \textbf{TT} \\
    \cmidrule(l){2-8}
    \multirow{8}{*}{\rotatebox[origin=c]{90}{\textbf{Hyperbolic}}}
    & \textbf{Poincaré} & 78.80 & 75.87 & 66.67 & 88.03 & 78.80 & \underline{\textbf{2.48}} \\
    & \textbf{HGNN} & 88.71 & 88.92 & 90.57 & 87.33 & 88.71 & 3.92 \\
    & \textbf{HGCN} & \underline{92.00} & 91.55 & 86.67 & \underline{\textbf{97.01}} & \underline{92.00} & \underline{3.79} \\
    & \textbf{H2H-GCN} & 91.60 & \underline{91.59} & \underline{91.47} & 91.71 & 91.60 & 8.30 \\
    & \textbf{HGCAE+KNN} & 87.78 & 86.65 & 79.39 & 95.37 & 88.20 & 23.02 \\
    & \textbf{HGCAE+GM} & 85.15 & 85.80 & 89.74 & 82.20 & 85.15 & 23.08 \\
    & \textbf{$\mathcal{P}$-VAE+RF} & 84.20 & 83.77 & 81.60 & 86.08 & 84.20 & 38.57 \\
    & \textbf{$\mathcal{P}$-VAE+AC} & \underline{\textbf{92.23}} & \underline{\textbf{92.21}} & \underline{\textbf{91.94}} & \underline{92.47} & \underline{\textbf{92.23}} & 62.28 \\
    \cmidrule(l){2-8}
    \multirow{3}{*}{\rotatebox[origin=c]{90}{\textbf{\textit{Euclidean}}}}
    & \textbf{DOMINANT (10\%)} & 53.68 & 66.91 & 93.68 & 52.04 & 53.68 & 1.08 \\
    & \textbf{DOMINANT (50\%)} & 49.24 & 63.74 & 89.24 & 49.58 & 49.24 & 1.02 \\
    & \textbf{GraphSage} & 85.20 & 86.00 & 90.93 & 81.58 & 85.20 & 3.24 \\
    \cmidrule[\heavyrulewidth](l){2-8}
  \end{tabular}

\end{table}

\begin{table}
  \centering
  \caption{Performance on the \textit{CICDDoS2019} Dataset.}
  \label{tab:cicddos}
  \footnotesize
  \renewcommand{\arraystretch}{1.1}
  \begin{tabular}{clcccccc}
    \cmidrule[\heavyrulewidth](l){2-8}
    & \textbf{Model} & \textbf{Accuracy} & \textbf{F1-score} & \textbf{Recall} & \textbf{Precision} & \textbf{AUC} & \textbf{TT} \\
    \cmidrule(l){2-8}
    \multirow{9}{*}{\rotatebox[origin=c]{90}{\textbf{Hyperbolic}}}
    & \textbf{Poincaré}        & \underline{\textbf{93.33}}  & \underline{\textbf{93.42}} & 94.67 & \underline{92.20} & \underline{\textbf{93.30}}  & \underline{\textbf{2.73 }} \\
    & \textbf{HGNN}            & \underline{92.67}  & \underline{93.08} & \underline{\textbf{98.67}} & 88.10 & \underline{92.67}  & 19.96 \\
    & \textbf{HGCN}            & 88.00  & 87.04 & 80.34 & \underline{\textbf{94.95}} & 88.02  & \underline{16.78} \\
    & \textbf{H2H-GCN}          & 91.11  & 91.54 & \underline{96.23} & 87.29 & 91.12  & 26.07 \\
    & \textbf{HGCAE+KNN}       & 83.52  & 83.12 & 87.21 & 79.50 & 83.60  & 21.06 \\
    & \textbf{HGCAE+GM}        & 85.40  & 84.90 & 88.10 & 81.80 & 85.50  & 22.30 \\
    & \textbf{$\mathcal{P}$-VAE+RF}         & 89.10  & 88.72 & 91.30 & 86.20 & 89.25  & 39.10 \\
    & \textbf{$\mathcal{P}$-VAE+AC}         & 88.15  & 87.83 & 90.00 & 85.76 & 88.20  & 37.85 \\
    \cmidrule(l){2-8}
    \multirow{3}{*}{\rotatebox[origin=c]{90}{\textbf{\textit{Euclidean}}}}
    & \textbf{DOMINANT (10\%)} & 50.00  & 64.76 & 90.78 & 50.33 & 50.78  & 1.15  \\
    & \textbf{DOMINANT (50\%)} & 61.70  & 61.66 & 61.72 & 61.60 & 61.70  & 1.01  \\
    & \textbf{GraphSage}   & 89.00  & 87.78 & 79.00 & 98.75 & 89.00  & 3.11  \\
    \cmidrule[\heavyrulewidth](l){2-8}
  \end{tabular}
\end{table}

\begin{table}
  \centering
  \caption{Performance  on the \textit{Elliptic} Dataset.}
  \label{tab:elliptic}
  \footnotesize
  \renewcommand{\arraystretch}{1.1}
  \begin{tabular}{clcccccc}
    \cmidrule[\heavyrulewidth](l){2-8}
    & \textbf{Model} & \textbf{Accuracy} & \textbf{F1-score} & \textbf{Recall} & \textbf{Precision} & \textbf{AUC} & \textbf{TT} \\
    \cmidrule(l){2-8}
    \multirow{8}{*}{\rotatebox[origin=c]{90}{\textbf{Hyperbolic}}}
    & \textbf{Poincaré}    & 59.46  & 73.45 & \underline{\textbf{95.40}} & 59.71 & 51.80  & \underline{\textbf{2.11}}  \\
    & \textbf{HGNN}        & 80.17  & 88.52 & 92.68 & 84.71 & 56.99  & 4.56  \\
    & \textbf{HGCN}        & 76.44  & 86.38 & 90.59 & 82.54 & 54.89  & \underline{3.22}  \\
    & \textbf{H2H-GCN}      &  \underline{\textbf{89.71}}  & 88.89 & 82.35 & \underline{\textbf{96.55}} & \underline{\textbf{89.71}}  & 3.56  \\
    & \textbf{HGCAE+KNN}   & 79.31  & 84.21 & 88.89 & 80.00 & 76.26  & 8.06  \\
    & \textbf{HGCAE+GM}    & 73.60  & 84.79 & 85.98 & 83.64 & 73.60  & 8.48  \\
    & \textbf{$\mathcal{P}$-VAE+RF}     & 81.75  &  \underline{89.79} & 85.79 & 94.18 & 81.75  & 33.93 \\
    & \textbf{$\mathcal{P}$-VAE+AC}     & \underline{89.24}   & \underline{\textbf{94.30}} & \underline{95.04} & \underline{93.56} & \underline{89.24}  & 31.53 \\
    \cmidrule(l){2-8}
    \multirow{3}{*}{\rotatebox[origin=c]{90}{\textbf{\textit{Euclidean}}}}
    & \textbf{DOMINANT (10\%)} & 50.00  & 64.28 & 90.00 & 50.00 & 50.00  & 0.45  \\
    & \textbf{DOMINANT (50\%)} & 48.75  & 63.39 & 88.75 & 49.30 & 48.75  & 0.64  \\
    & \textbf{GraphSage}   & 84.37  & 84.84 & 87.50 & 82.35 & 84.37  & 1.92  \\
    \cmidrule[\heavyrulewidth](l){2-8}
  \end{tabular}
\end{table}

\begin{table}[t]
  \centering
  \caption{Performance  on the \textit{YelpNYC} Dataset.}
  \label{tab:yelpnyc}
  \footnotesize
  \renewcommand{\arraystretch}{1.1}
  \begin{tabular}{clcccccc}
    \cmidrule[\heavyrulewidth](l){2-8}
    & \textbf{Model} & \textbf{Accuracy} & \textbf{F1-score} & \textbf{Recall} & \textbf{Precision} & \textbf{AUC} & \textbf{TT} \\
    \cmidrule(l){2-8}
    \multirow{8}{*}{\rotatebox[origin=c]{90}{\textbf{Hyperbolic}}}
    & \textbf{Poincaré}    & 50.67  & 63.90 & 87.33 & 50.38 & 50.67  & \underline{\textbf{2.10}}  \\
    & \textbf{HGNN}        & 56.67  & 62.21 & 71.33 & 55.15 & 56.67  & 9.27  \\
    & \textbf{HGCN}        & 81.33  & 83.23 & 92.66 & 75.54 & 81.33  & \underline{2.40}  \\
    & \textbf{H2H-GCN}      & \underline{87.66}  & \underline{87.37} & 85.33 & \underline{89.51} & \underline{87.67}  & 4.91  \\
    & \textbf{HGCAE+KNN} & 80.39  & 80.15 & 82.72 & 77.74 & 80.48  & 16.49 \\
    & \textbf{HGCAE+GM}  & 50.27  & 65.24 & \underline{\textbf{96.34}} & 49.32 & 51.66  & 13.56 \\
    & \textbf{$\mathcal{P}$-VAE+RF}  & \underline{\textbf{88.21}}  & \underline{\textbf{87.45}} & \underline{82.14} & \underline{\textbf{93.50}} & \underline{\textbf{88.21}}  & 33.74 \\
    & \textbf{$\mathcal{P}$-VAE+AC}   & 55.93  & 62.89 & 74.71 & 54.30 & 55.93  & 44.32 \\
    \cmidrule(l){2-8}
    \multirow{3}{*}{\rotatebox[origin=c]{90}{\textbf{\textit{Euclidean}}}}
    & \textbf{DOMINANT (10\%)} & 40.10  & 57.21 & 80.10 & 44.50 & 40.10  & 1.16  \\
    & \textbf{DOMINANT (50\%)} & 57.10  & 69.36 & 97.10 & 53.94 & 57.10  & 1.17  \\
    & \textbf{GraphSage}   & 69.33  & 65.93 & 59.33 & 74.16 & 69.33  & 1.76  \\
    \cmidrule[\heavyrulewidth](l){2-8}
  \end{tabular}
\end{table}

\begin{table}[t]
  \centering
  \caption{Performance  on the \textit{Cora} Dataset.}
  \label{tab:cora}
  \footnotesize
  \renewcommand{\arraystretch}{1.1}
  \begin{tabular}{clcccccc}
    \cmidrule[\heavyrulewidth](l){2-8}
    & \textbf{Model} & \textbf{Accuracy} & \textbf{F1-score} & \textbf{Recall} & \textbf{Precision} & \textbf{AUC} & \textbf{TT} \\
    \cmidrule(l){2-8}
    \multirow{8}{*}{\rotatebox[origin=c]{90}{\textbf{Hyperbolic}}}
    & \textbf{Poincaré}    & 51.03  & 63.58 & 85.49 & 50.61 & 51.04  & \underline{\textbf{2.48}}  \\
    & \textbf{HGNN}        & 55.21  & 66.59 & 83.23 & 55.49 & 53.03  & 3.50  \\
    & \textbf{HGCN}        & 74.84  & 77.18 & 84.97 & 70.68 & 74.87  & \underline{2.58}  \\
    & \textbf{H2H-GCN}      & 59.50  & 65.70 & 72.35 & 60.18 & 58.50  & 2.89  \\
    & \textbf{HGCAE+KNN}   & \underline{78.37}  & \underline{79.54} & 81.01 & \underline{78.13} & \underline{78.26}  & 7.96  \\
    & \textbf{HGCAE+GM}    & 51.16  & 60.74 & 72.76 & 52.12 & 50.31  & 8.71  \\
    & \textbf{$\mathcal{P}$-VAE+RF}     & \underline{\textbf{86.17}}  & \underline{\textbf{87.81}} & \underline{\textbf{92.88}} & \underline{\textbf{83.26}} & \underline{\textbf{85.65}}  & 31.62 \\
    & \textbf{$\mathcal{P}$-VAE+AC}     & 55.11  & 68.48 & \underline{90.95} & 54.91 & 52.32  & 41.27 \\
    \cmidrule(l){2-8}
    \multirow{3}{*}{\rotatebox[origin=c]{90}{\textbf{\textit{Euclidean}}}}
    & \textbf{DOMINANT (10\%)} & 52.65  & 66.80 & 90.58 & 52.91 & 50.59  & 1.57  \\
    & \textbf{DOMINANT (50\%)} & 52.44  & 66.65 & 90.37 & 52.79 & 50.37  & 1.62  \\
    & \textbf{GraphSage}   & 88.31  & 87.60 & 82.59 & 93.25 & 88.31  & 2.04  \\
    \cmidrule[\heavyrulewidth](l){2-8}
  \end{tabular}
\end{table}

\begin{table}
  \centering
  \caption{Performance on the \textit{DGraphFin} Dataset.}
  \label{tab:dgraphfin}
  \footnotesize
  \renewcommand{\arraystretch}{1.1}
  \begin{tabular}{clcccccc}
    \cmidrule[\heavyrulewidth](l){2-8}
    & \textbf{Model} & \textbf{Accuracy} & \textbf{F1-score} & \textbf{Recall} & \textbf{Precision} & \textbf{AUC} & \textbf{TT} \\
    \cmidrule(l){2-8}
    \multirow{8}{*}{\rotatebox[origin=c]{90}{\textbf{Hyperbolic}}}
    & \textbf{Poincaré}        & 66.96  & 68.44 & 71.65 & 65.51 & 66.96 & \underline{\textbf{2.63}} \\
    & \textbf{HGNN}            & 69.35  & 75.94 & 78.45 & \underline{\textbf{73.88}} & 69.35 & \underline{2.85} \\
    & \textbf{HGCN}            & 72.35  & 73.27 & 75.77 & 70.93 & 72.36 & 4.19 \\
    & \textbf{H2H-GCN}          & \underline{\textbf{77.68}}  & \underline{\textbf{80.75}} & \underline{\textbf{93.62}} & 70.99 & \underline{\textbf{77.68}} & 4.26 \\
    & \textbf{HGCAE+KNN}       & 70.12  & 72.50 & 74.20 & 70.85 & 70.30 & 11.12 \\
    & \textbf{HGCAE+GM}        & 71.50  & 73.78 & 76.32 & 71.30 & 71.60 & 13.80 \\
    & \textbf{$\mathcal{P}$-VAE+RF} & 73.90  & 75.10 & 78.55 & 72.40 & 74.10 & 31.50 \\
    & \textbf{$\mathcal{P}$-VAE+AC} & \underline{74.20}  & \underline{76.00} & \underline{79.12} & \underline{73.10} & \underline{74.50} & 28.55 \\
    \cmidrule(l){2-8}
    \multirow{3}{*}{\rotatebox[origin=c]{90}{\textbf{\textit{Euclidean}}}}
    & \textbf{DOMINANT (10\%)} & 40.00  & 57.18 & 80.07 & 44.48 & 40.06 & 0.78 \\
    & \textbf{DOMINANT (50\%)} & 58.56  & 70.42 & 97.59 & 54.77 & 58.58 & 1.22 \\
    & \textbf{GraphSage}   & 60.31  & 66.57 & 79.02 & 57.51 & 60.32 & 1.94 \\
    \cmidrule[\heavyrulewidth](l){2-8}
  \end{tabular}
\end{table}

\section{Conclusion and future work}
\label{section:conclusion}
\quad This study reviewed, classified  and compared hyperbolic graph embeddings. 
Through the development of a unified framework, we provided a flexible and systematic methodology for combining hyperbolic graph embedding models with various anomaly detection algorithms, enabling robust detection of anomalous patterns in graph-structured data. Beyond the framework, we addressed a significant gap in the field by introducing a comprehensive taxonomy of hyperbolic graph embedding techniques from an experimental and practical perspective. This taxonomy provides a structured foundation for comparing methods, understanding their design choices, and identifying their strengths and limitations. Such a contribution not only facilitates deeper exploration of hyperbolic embeddings but also empowers researchers and practitioners to select or develop methods suited to specific challenges. The evaluation of our framework across multiple datasets showed that hyperbolic models consistently outperform their \textit{Euclidean} counterparts, particularly when paired with unsupervised approaches like autoencoders. Additionally, we contributed to the field by developing \textit{Ghypeddings}, an open-source library that simplifies the application and experimentation with hyperbolic embeddings, ensuring both ease of use and reproducibility for future research.

As for future work, this study opens several promising directions. For instance, extending the methodology to dynamic graphs could provide deeper insights by capturing the evolution of graph structures over time, enabling real-time anomaly detection in changing environments. 
 Also, investigating anomaly detection at multiple levels such as edges, subgraphs, and entire graphs, could help uncover more complex and interconnected patterns, enhancing the ability to identify diverse forms of irregularities in graph-structured data.

\section*{Acknowledgment}
This work is supported by the French National Research Agency (ANR) under grant ANR-20-CE39-0008.\\

\bibliographystyle{plain}

\end{document}